\documentclass[letterpaper]{article}
\usepackage{nips07submit_e,times}
\usepackage{times}
\usepackage{amsmath}
\usepackage{amssymb}
\usepackage{amsthm}
\usepackage{amsfonts}
\usepackage{graphicx}
\usepackage{graphicx}\graphicspath{{figures/}}
\usepackage{subfigure}

\usepackage{epsfig}
\usepackage{graphicx}
\usepackage{algorithm}
\usepackage{algorithmic}
\usepackage{amsthm}

\title{Bayesian Agglomerative Clustering with Coalescents}

\author{
Yee Whye Teh \\
Gatsby Unit \\
University College London\\
{\tt ywteh@gatsby.ucl.ac.uk}
\And
Hal Daum\'e III \\
School of Computing \\
University of Utah\\
{\tt me@hal3.name}
\And
Daniel Roy \\
CSAIL\\
MIT\\
{\tt droy@mit.edu}
}

\DeclareMathOperator*{\Exp}{Exp}

\newcommand{\secref}[1]{Section~\ref{sec:#1}}
\newcommand{\figref}[1]{Figure~\ref{fig:#1}}

\renewcommand{\eqref}[1]{(\ref{eq:#1})}

\def\hy{{\widehat{y}}}

\def\vx{{\mathbf{x}}}
\def\vy{{\mathbf{y}}}

\def\bone{\pmb{1}}

\newcommand{\La}{\Lambda}
\newcommand{\la}{\lambda}

\def\T{^\top}
\newcommand{\inv}{^{-1}}
\newcommand{\norm}[1]{\left\vert\left\vert #1 \right\vert\right\vert}
\newcommand{\nchooser}[2]{\big(\!\begin{smallmatrix} #1\\#2\end{smallmatrix}\!\big)}
\newcommand{\alg}[1]{\textsf{\small #1}}

\newcommand{\halcomment}[1]{}

\renewcommand\ldots{.\hspace*{.15ex}.\hspace*{.15ex}.\hspace*{.15ex}}
\renewcommand\cdots{\cdot\hspace*{-.25ex}\cdot\hspace*{-.25ex}\cdot\hspace*{-.25ex}}

\begin{document}
\maketitle

\begin{abstract}
We introduce a new Bayesian model for hierarchical clustering based on a prior over trees called Kingman's coalescent.  We develop novel greedy and sequential Monte Carlo inferences which operate in a bottom-up agglomerative fashion.  We show experimentally the superiority of our algorithms over others, and demonstrate our approach in document clustering and phylolinguistics.
\end{abstract}

\section{Introduction}

Hierarchically structured data abound across a wide variety of domains.  It is thus not surprising that hierarchical clustering is a traditional mainstay of machine learning \cite{DudHar1973}.  The dominant approach to hierarchical clustering is agglomerative: start with one cluster per datum, and greedily merge pairs until a single cluster remains.  Such algorithms are efficient and easy to implement.  Their primary limitations---a lack of predictive semantics and a coherent mechanism to deal with missing data---can be addressed by probabilistic models that handle partially observed data, quantify goodness-of-fit, predict on new data, and integrate within more complex models, all in a principled fashion.


Currently there are two main approaches to probabilistic models for hierarchical clustering.  The first takes a direct Bayesian approach by defining a prior over trees followed by a distribution over data points conditioned on a tree \cite{Nea2001,Wil2000,KemGriStr2004,RoyKemMan2007}.  MCMC sampling is then used to obtain trees from their posterior distribution given observations.  This approach has the advantages and disadvantages of most Bayesian models: averaging over sampled trees can improve predictive capabilities, give confidence estimates for conclusions drawn from the hierarchy, and share statistical strength across the model; but it is also computationally demanding and complex to implement.  As a result such models have not found widespread use.  \cite{Nea2001} has the additional advantage that the distribution induced on the data points is exchangeable, so the model can be coherently extended to new data.  The second approach uses a flat mixture model as the underlying probabilistic model and structures the posterior hierarchically \cite{HelGha2005,Fri2003}.  This approach uses an agglomerative procedure to find the tree giving the best posterior approximation, mirroring traditional agglomerative clustering techniques closely and giving efficient and easy to implement algorithms.  However because the underlying model has no hierarchical structure, there is no sharing of information across the tree.

We propose a novel class of Bayesian hierarchical clustering models and associated inference algorithms combining the advantages of both probabilistic approaches above.  1) We define a prior and compute the posterior over trees, thus reaping the benefits of a fully Bayesian approach; 2) the distribution over data is hierarchically structured allowing for sharing of statistical strength; 3) we have efficient and easy to implement inference algorithms that construct trees agglomeratively; and 4) the induced distribution over data points is exchangeable.  
Our model is based on an exchangeable distribution over trees called Kingman's coalescent \cite{Kin1982a,Kin1982b}.  Kingman's coalescent is a standard model from population genetics for the genealogy of a set of individuals.  It is obtained by tracing the genealogy backwards in time, noting when lineages \emph{coalesce} together.  We review Kingman's coalescent in \secref{coalescent}.  Our own contribution is in using it as a prior over trees in a hierarchical clustering model (\secref{bac}) and in developing novel inference procedures for this model (\secref{inference}).

\halcomment{setup
$P(\pi,X)=P(\pi)P(X|\pi)$
next section concentrate on $P(\pi)$, the next on $P(X|\pi)$, then approximate inference, then experiments
}
\section{Kingman's coalescent}\label{sec:coalescent}

\begin{figure}[t]
\includegraphics[width=\linewidth]{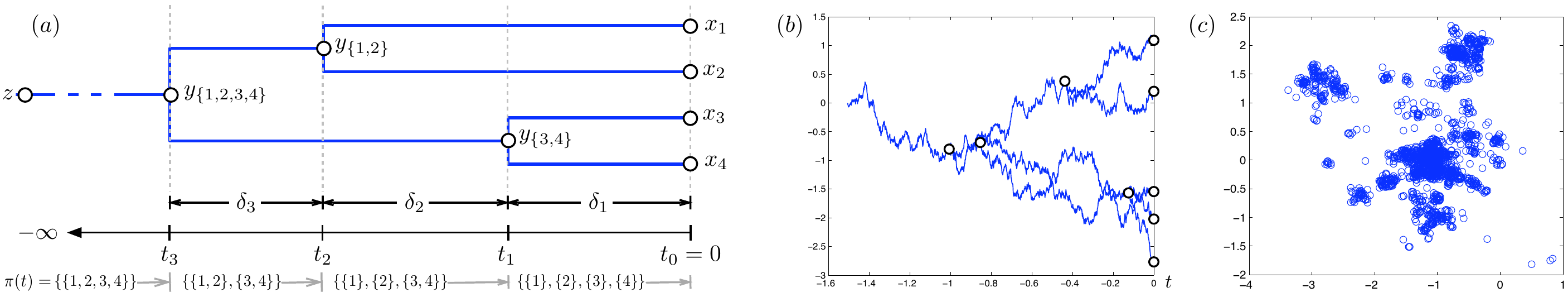}
\caption{(a) Variables describing the $n$-coalescent. (b) Sample path from a Brownian diffusion coalescent process in 1D, circles are coalescent points. (c) Sample observed points from same in 2D, notice the hierarchically clustered nature of the points.}
\label{fig:coalescent}
\end{figure}

Kingman's coalescent is a standard model in population genetics describing the common genealogy (ancestral tree) of a set of individuals \cite{Kin1982a,Kin1982b}.  In its full form it is a distribution over the genealogy of a countably infinite set of individuals.  Like other nonparametric models (e.g.\ Gaussian and Dirichlet processes), Kingman's coalescent is most easily described and understood in terms of its finite dimensional marginal distributions over the genealogies of $n$ individuals, called $n$-coalescents. We obtain Kingman's coalescent as $n\!\rightarrow\!\infty$.

Consider the genealogy of $n$ individuals alive at the present time $t=0$.  We can trace their ancestry backwards in time to the distant past $t\!=\!-\!\infty$.  Assume each individual has one parent (in genetics, \emph{haploid} organisms), and therefore genealogies of $[n]=\{1,\ldots,n\}$ form a \emph{directed forest}.   In general, at time $t\!\le\! 0$, there are $m$ ($1\!\le\! m\! \le\! n$) ancestors alive.  Identify these ancestors with their corresponding sets $\rho_{1},\ldots,\rho_{m}$ of descendants (we will make this identification throughout the paper).  Note that $\pi(t)=\{\rho_{1},\ldots,\rho_{m}\}$ form a \emph{partition} of $[n]$, and interpret $t\!\mapsto\! \pi(t)$ as a function from $(-\!\infty,0]$ to the set of partitions of $[n]$.  This function is piecewise constant, left-continuous, monotonic ($s\!\le\! t$ implies that $\pi(t)$ is a refinement of $\pi(s)$), and $\pi(0)\!=\!\{\{1\},\ldots,\{n\}\}$ (see \figref{coalescent}a).  Further, $\pi$ \emph{completely and succinctly} characterizes the genealogy; we shall henceforth refer to $\pi$ as \emph{the genealogy} of $[n]$.  


Kingman's $n$-coalescent is simply a distribution over genealogies of $[n]$, or equivalently, over the space of partition-valued functions like $\pi$.  More specifically, the $n$-coalescent is a continuous-time, partition-valued, Markov process, which starts at $\{\{1\},\ldots,\{n\}\}$ at present time $t\!=\!0$, and evolves \emph{backwards in time}, merging (coalescing) lineages until only one is left.  To describe the Markov process in its entirety, it is sufficient to describe the jump process (i.e.\ the embedded, discrete-time, Markov chain over partitions) and the distribution over coalescent times.  Both are straightforward and their simplicity is part of the appeal of Kingman's coalescent.  Let $\rho_{li},\rho_{ri}$ be the $i$th pair of lineages to coalesce, $t_{n-1}\!<\!\cdots\!<\!t_{1}\!<\!t_0\!=\!0$ be the coalescent times and $\delta_i \!=\! t_{i-1}\! -\! t_i \!>\!0$ be the duration between adjacent events (see \figref{coalescent}a).  Under the $n$-coalescent, every pair of lineages merges independently with rate 1.  Thus the first pair amongst $m$ lineages merge with rate $\nchooser{m}2 \!=\! \frac{m(m-1)}2$.  Therefore $\delta_i\!\sim\!\Exp\left(\nchooser{n-i+1}{2}\right)$ independently, the pair $\rho_{li},\rho_{ri}$ is chosen from among those right after time $t_i$, and with probability one a random draw from the $n$-coalescent is a binary tree with a single root at $t\!=\!-\!\infty$ and the $n$ individuals at time $t\!=\!0$.
The genealogy is given as:
\begin{align}
\pi(t) = \begin{cases} 
\{\{1\},\ldots,\{n\}\}  & \text{if $t=0$;} \\
\pi_{t_{i-1}}-\rho_{li}-\rho_{ri} + (\rho_{li} \cup \rho_{ri})   & \text{if $t=t_i$;} \\
\pi_{t_i}  & \text{if $t_{i+1}<t<t_i$.}
\end{cases}
\end{align}
Combining the probabilities of the durations and choices of lineages, the probability of $\pi$ is simply:
\begin{align}
p(\pi) &=\textstyle \prod_{i=1}^{n-1} \nchooser{n-i+1}{2}\exp\left(-\nchooser{n-i+1}{2}\delta_i\right) / \nchooser{n-i+1}{2} = \prod_{i=1}^{n-1}  \exp\left(-\nchooser{n-i+1}{2}\delta_i\right)
\label{eq:prior}
\end{align}
The $n$-coalescent has some interesting statistical properties \cite{Kin1982a,Kin1982b}.  The marginal distribution over tree topologies is uniform and independent of the coalescent times.  Secondly, it is infinitely exchangeable: given a genealogy drawn from an $n$-coalescent, the genealogy of any $m$ contemporary individuals alive at time $t\!\le\!0$ embedded within the genealogy is a draw from the $m$-coalescent. Thus, taking $n\!\rightarrow\!\infty$, there is a distribution over genealogies of a countably infinite population for which the marginal distribution of the genealogy of any $n$ individuals gives the $n$-coalescent.  Kingman called this \emph{the coalescent}.  

\section{Hierarchical clustering with coalescents}\label{sec:bac}

We take a Bayesian approach to hierarchical clustering, placing a coalescent prior on the latent tree and modeling observed data with a Markov process evolving \emph{forward in time} along the tree.  We will alter our terminology from genealogy to tree, from $n$ individuals at present time to $n$ observed data points, and from individuals on the genealogy to latent variables on the tree-structured distribution.  Let $x_1,\ldots,x_n$ be $n$ observed data at the leaves of a tree $\pi$ drawn from the $n$-coalescent.  $\pi$ has $n-1$ coalescent points, the $i$th occuring when $\rho_{li}$ and $\rho_{ri}$ merge at time $t_i$ to form $\rho_i=\rho_{li}\cup\rho_{ri}$.  Let $t_{li}$ and $t_{ri}$ be the times at which  $\rho_{li}$ and $\rho_{ri}$ are themselves formed.  

We construct a continuous-time Markov process evolving along the tree from the past to the present, branching independently at each coalescent point until we reach time 0, where the $n$ Markov processes induce a distribution over the $n$ data points.  The joint distribution respects the conditional independences implied by the structure of the directed tree.  Let $y_{\rho_i}$ be a latent variable that takes on the value of the Markov process at $\rho_i$ just before it branches (see \figref{coalescent}a).  Let $y_{\{i\}}=x_i$ at leaf $i$. 


To complete the description of the likelihood model, let $q(z)$ be the initial distribution of the Markov process at time $t=-\infty$, and $k_{st}(x,y)$ be the transition probability from state $x$ at time $s$ to state $y$ at time $t$.  This Markov process need be neither stationary nor ergodic.  Marginalizing over paths of the Markov process, the joint probability over the latent variables and the observations is:
\begin{equation} \textstyle
p(\vx,\vy,z|\pi)
= q(z)k_{-\infty\,t_{n-1}}(z,y_{\rho_{n-1}}) 
\prod_{i=1}^{n-1} k_{t_i t_{li}}(y_{\rho_i},y_{\rho_{li}})k_{t_i t_{ri}}(y_{\rho_i},y_{\rho_{ri}})
\label{eq:probxyz}
\end{equation}
Notice that the marginal distributions at each observation $p(x_i|\pi)$ are identical and given by the Markov process at time $0$.  However, they are not independent: they share the same sample path down the Markov process until they split.  In fact the amount of dependence between two observations is a function of the time at which the observations coalesce in the past.  A more recent coalescent time implies larger dependence.  \halcomment{In other words, the process induces a hierarchical structure on the joint distribution $p(\vx|\pi)$, such that data points closer in the tree are more correlated.} The overall distribution induced on the observations $p(\vx)$ inherits the infinite exchangeability of the $n$-coalescent.  We considered a brownian diffusion (see Figures~\ref{fig:coalescent}(b,c)) and a simple independent sites mutation process on multinomial vectors (\secref{examples}).

\section{Agglomerative sequential Monte Carlo and greedy inference}\label{sec:inference}

We develop two classes of efficient and easily implementable inference algorithms for our hierarchical clustering model based on sequential Monte Carlo (SMC) and greedy schemes respectively.  In both classes, the latent variables are integrated out, and the trees are constructed in a bottom-up fashion.  The full tree $\pi$ can be expressed as a series of $n-1$ coalescent events, ordered backwards in time.  The $i$th coalescent event involves the merging of the two subtrees with leaves $\rho_{li}$ and $\rho_{ri}$ and occurs at a time $\delta_i$ before the previous coalescent event.  Let $\theta_i=\{\delta_j,\rho_{lj},\rho_{rj}\text{ for $j\le i$}\}$ denote the first $i$ coalescent events. \halcomment{ equivalently the partially constructed tree after the first $i$ iterations of the algorithms.}  $\theta_{n-1}$ is equivalent to $\pi$ and we shall use them interchangeably.

We assume that the form of the Markov process is such that the latent variables $\{y_{\rho_i}\}_{i=1}^{n-1}$ and $z$ can be efficiently integrated out using an upward pass of belief propagation on the tree.  Let $M_{\rho_i}(y)$ be the message passed from $y_{\rho_i}$ to its parent; $M_{\{i\}}(y) = \delta_{x_i}(y)$ is point mass at $x_i$ for leaf $i$.  $M_{\rho_i}(y)$ is proportional to the likelihood of the observations at the leaves below coalescent event $i$, given that $y_{\rho_i} = y$.  Belief propagation computes the messages recursively up the tree; for $i=1,\ldots,n-1$:
\begin{align} \textstyle
M_{\rho_i}(y) = {Z^{-1}_{\rho_i}(\vx,\theta_{i})}\prod_{b=l,r}
\int k_{t_it_{bi}}(y,y_b)M_{\rho_{bi}}(y_b)\,dy_b
\end{align}
$Z_{\rho_i}(\vx,\theta_{i})$ is a normalization constant introduced to avoid numerical problems.  \halcomment{, and is a function only of the data $\vx$ and the first $i$ coalescent events $\theta_i$ (importantly it does not depend on the coalescent events after the $i$th one).}  The choice of $Z$ does not affect the probability of $\vx$, but does impact the accuracy and efficiency of our inference algorithms.  We found that $Z_{\rho_i}(\vx,\theta_i) = \int q(y) M_{\rho_i}(y)\, dy$ worked well.
At the root, we have:
\begin{equation} \textstyle
Z_{-\infty}(\vx,\theta_{n-1}) = \int q(z)k_{-\infty\,t_{n-1}}(z,y) M_{\rho_{n-1}}(y) \,dy\,dz \label{eq:top}
\end{equation}
The marginal probability $p(\vx|\pi)$ is now given by the product of normalization constants:
\begin{align} \textstyle
p(\vx|\pi) = Z_{-\infty}(\vx,\theta_{n-1})\prod_{i=1}^{n-1} Z_{\rho_i}(\vx,\theta_{i})\label{eq:probx}
\end{align}
Multiplying in the prior \eqref{prior} over $\pi$, we get the joint probability for the tree $\pi$ and observations $\vx$:
\begin{equation} \textstyle
p(\vx,\pi) = Z_{-\infty}(\vx,\theta_{n-1})\prod_{i=1}^{n-1} 
\exp\left(-\nchooser{n-i+1}{2}\delta_i\right)Z_{\rho_i}(\vx,\theta_{i})
\label{eq:joint}
\end{equation}
Our inference algorithms are based upon \eqref{joint}.  Note that each term $Z_{\rho_i}(\vx,\theta_i)$ can be interpreted as a local likelihood term for coalescing the pair $\rho_{li}$, $\rho_{ri}$\footnote{If the Markov process is stationary with equilibrium $q(y)$, $Z_{\rho_i}(\vx,\theta_i)$ is a likelihood ratio between two models with observations $\vx_{\rho_i}$: (1) a single tree with leaves $\rho_i$; (2) two independent trees with leaves $\rho_{li}$ and $\rho_{ri}$ respectively.  This is similar to \cite{HelGha2005,Fri2003} and is used later in our NIPS experiment to determine coherent clusters.}.  
In general, for each $i$, we choose a duration $\delta_i$ and a pair of subtrees $\rho_{li}$, $\rho_{ri}$ to coalesce.  This choice is based upon the $i$th term in \eqref{joint}, interpreted as the product of a local prior and a local likelihood for choosing $\delta_i$, $\rho_{li}$ and $\rho_{ri}$ given $\theta_{i-1}$.  \halcomment{If the duration and subtrees are picked to optimize the $i$th term or some surrogate, we obtain a greedy algorithm, while if they are sampled according to proposal distributions and reweighted using the $i$th term appropriately, we obtain a SMC algorithm.}

\halcomment{
The specific choice of the type of normalization used does not matter to the computation of the probability of $\vx$ given $\pi$ in \eqref{probx}.  However this choice does affect the performance of the inference algorithms---a bad choice will increase the variance of the SMC samplers and decrease the quality of the greedily constructed trees.  In our experience using Brownian diffusion and the Markov process on binary vectors, we have found the following to work well.  Note that both Markov process are stationary, reversible and mixing, thus there is an equilibrium distribution $q(z)$ and a stationary kernel $k_{st}(y,z)=k_{|s-t|}(y,z)$ such that $q(z)k_{|s-t|}(z,y)=k_{|s-t|}(y,z)q(y)$.  We used:
\begin{align}
Z_{\rho_i}(\vx,\theta_i) &= \int q(y) \prod_{b=l,r} \int k_{t_it_{bi}}(y,y_b)M_{\rho_{bi}}(y_b)\,dy_b\,dy \nonumber \\
&= \int\int q(y_l) k_{t_{li}+t_{ri}-2t_i}(y_l,y_r) M_{\rho_{li}}(y_l)M_{\rho_{ri}}(y_r)\,dy_l\,dy_r \label{eq:locallik}
\end{align}
}
\halcomment{
$Z_{\rho_i}$'s are likelihood ratios of merging vs not merging.

Partially observed data

Predictive probabilities

Hyperparameter optimization

Computational costs.  

Solving for $\delta_i$.
}

\subsection{Sequential Monte Carlo algorithms}
\def\ss{^{s}}
Sequential Monte Carlo algorithms (aka particle filters), approximate the posterior using a weighted sum of point masses \cite{SMCbook}.  These point masses are constructed iteratively.  At iteration $i-1$, particle $s$ consists of $\theta_{i-1}\ss=\{\delta_j\ss, \rho_{lj}\ss, \rho_{rj}\ss \text{ for $j<i$}\}$, and has weight $w_{i-1}\ss$.  At iteration $i$, $s$ is extended by sampling $\delta_i\ss$, $\rho_{li}\ss$ and $\rho_{ri}\ss$ from a proposal distribution $f_i(\delta_i\ss,\rho_{li}\ss,\rho_{ri}\ss|\theta_{i-1}\ss)$, with weights:
\begin{align}
w_i\ss = w_{i-1}\ss {\exp\left(-\nchooser{n-i+1}{2}\delta_i\ss\right)Z_{\rho_i}(\vx,\theta_{i}\ss)}/{f_i(\delta_i\ss,\rho_{li}\ss,\rho_{ri}\ss|\theta_{i-1}\ss)} \label{eq:smcweight}
\end{align}
After $n-1$ iterations, we obtain a set of trees $\theta_{n-1}\ss$ and weights $w_{n-1}\ss$.  The joint distribution is approximated by:
$p(\pi,\vx) \approx \sum_s w_{n-1}\ss \delta_{\theta_{n-1}\ss}(\pi)$,
while the posterior is approximated with the weights normalized.  An important aspect of SMC is resampling, which places more particles in high probability regions and prunes particles stuck in low probability regions.  We resample as in Algorithm 5.1 of \cite{Fea1998} when the effective sample size ratio as estimated in \cite{Nea1998a} falls below one half.

\textbf{\alg{SMC-PriorPrior}}.  The simplest proposal distribution is to sample $\delta_i\ss$, $\rho_{li}\ss$ and $\rho_{ri}\ss$ from the local prior.  $\delta_i\ss$ is drawn from an exponential with rate $\nchooser{n-i+1}{2}$ and $\rho_{li}\ss, \rho_{ri}\ss$ are drawn uniformly from all available pairs.  The weight updates \eqref{smcweight} reduce to multiplying by $Z_{\rho_i}(\vx,\theta_i\ss)$.  This approach is computationally very efficient, but performs badly with many objects due to the uniform draws over pairs.\halcomment{  Specifically, the pairs are drawn uniformly, while instead we should pick pairs that are closer (in terms of higher $Z_{\rho_i}$) more often.}
\textbf{\alg{SMC-PriorPost}}.  The second approach addresses the suboptimal choice of pairs to coalesce.  We first draw $\delta_i\ss$ from its local prior, then draw $\rho_{li}\ss$, $\rho_{ri}\ss$ from the local posterior:
\begin{equation} \textstyle
f_i(\rho_{li}\ss,\rho_{ri}\ss|\delta_i\ss,\theta_{i-1}\ss) 
\propto \textstyle {Z_{\rho_i}(\vx,\theta_{i-1}\ss,\delta_i\ss,\rho_{li}\ss,\rho_{ri}\ss)}
\halcomment{{\sum_{\rho_{l}',\rho_{r}'}Z_{\rho_i}(\vx,\theta_{i-1}\ss,\delta_i\ss,\rho_{l}',\rho_{r}')}}; \quad
w_i\ss = w_{i-1}\ss\sum_{\rho_{l}',\rho_{r}'}Z_{\rho_i}(\vx,\theta_{i-1}\ss,\delta_i\ss,\rho_{l}',\rho_{r}')
\end{equation}
This approach is more computationally demanding since we need to evaluate the local likelihood of every pair.  It also performs significantly better than \alg{SMC-PriorPrior}.  We have found that it works reasonably well for small data sets but fails in larger ones for which the local posterior for $\delta_i$ is highly peaked.
\textbf{\alg{SMC-PostPost}}. The third approach is to draw all of $\delta_i\ss$, $\rho_{li}\ss$ and $\rho_{ri}\ss$ from their posterior:
\begin{align}
f_i(\delta_i\ss,\rho_{li}\ss,\rho_{ri}\ss|\theta_{i-1}\ss) 
&\propto \textstyle {\exp\left(-\nchooser{n-i+1}{2}\delta_i\ss\right)Z_{\rho_i}(\vx,\theta_{i-1}\ss,\delta_i\ss,\rho_{li}\ss,\rho_{ri}\ss)}
\halcomment{{\sum_{\rho_{l}',\rho_{r}'}\int \exp\left(-\nchooser{n-i+1}{2}\delta'\right)Z_{\rho_i}(\vx,\theta_{i-1}\ss,\delta',\rho_{l}',\rho_{r}')\,d\delta'}} \nonumber \\
w_i\ss &= \textstyle w_{i-1}\ss\sum_{\rho_{l}',\rho_{r}'}\int\exp\left(-\nchooser{n-i+1}{2}\delta'\right)Z_{\rho_i}(\vx,\theta_{i-1}\ss,\delta',\rho_{l}',\rho_{r}') \,d\delta'
\end{align}
This approach requires the fewest particles, but is the most computationally expensive due to the integral for each pair.  Fortunately, for the case of Brownian diffusion process described below, these integrals are tractable and related to generalized inverse Gaussian distributions.

\subsection{Greedy algorithms}

SMC algorithms are attractive because they produce an arbitrarily accurate approximation to the full posterior.  However in many applications a single good tree is often times sufficient.  We describe a few greedy algorithms to construct a good tree.

\textbf{\alg{Greedy-MaxProb}}: the obvious greedy algorithm is to pick $\delta_i$, $\rho_{li}$ and $\rho_{ri}$ maximizing the $i$th term in \eqref{joint}.  We do so by computing the optimal $\delta_i$ for each pair of $\rho_{li}$, $\rho_{ri}$, and then picking the pair maximizing the $i$th term at its optimal $\delta_i$. \halcomment{Interestingly, \alg{Greedy-MaxProb} need not choose a pair with the smallest optimal duration.  Thus, there could be pairs for which the optimal duration maximizing the $i$th term in \eqref{joint} is negative.  If this pair were chosen we simply allow it to coalesce at time $t_{i-1}+\delta_i$ regardless, adjusting the time ordered list of coalescent events accordingly.  }
\textbf{\alg{Greedy-MinDuration}}: simply pick the pair to coalesce whose optimal duration is minimum.  
Both algorithms require recomputing the optimal duration for each pair at each iteration, since the exponential rate $\nchooser{n-i+1}{2}$ on the duration varies with the iteration $i$.  The total computational cost is thus $O(n^3)$.  We can avoid this by using the alternative view of the $n$-coalesent as a Markov process where each pair of lineages coalesces at rate 1.
\textbf{\alg{Greedy-Rate1}}: for each pair $\rho_{li}$ and $\rho_{ri}$ we determine the optimal $\delta_i$, but replacing the $\nchooser{n-i+1}{2}$ prior rate with 1.  We coalesce the pair with most recent time (as in \alg{Greedy-MinDuration}).  This reduces the complexity to $O(n^2)$.  We found that all three perform about equally well.

\halcomment{\subsection{Other considerations} }

\subsection{Examples}\label{sec:examples}

\textbf{Brownian diffusion}. Consider the case of continuous data evolving via Brownian diffusion.  The transition kernel $k_{st}(y,\cdot)$ is a Gaussian centred at $y$ with variance $(t-s)\Lambda$, where $\Lambda$ is a symmetric p.d.\ covariance matrix.  Because the joint distribution \eqref{probxyz} over $\vx$, $\vy$ and $z$ is Gaussian, we can express each message $M_{\rho_i}(y)$ as a Gaussian with mean $\hy_{\rho_i}$ and variance $\Lambda v_{\rho_i}$.  The local likelihood is:
\begin{align}
Z_{\rho_i}(\vx,\theta_i) &= \textstyle
|2\pi\widehat \Lambda_i|^{-\frac{1}{2}} 
\exp\bigl( -\textstyle\frac 1 2 
\norm{\hy_{\rho_{li}}\!-\!\hy_{\rho_{ri}}}^2_{\widehat \Lambda_i}
\bigr);&
\widehat\Lambda_i &= \Lambda(v_{\rho_{li}}\!+\!v_{\rho_{ri}}\!+\!t_{li}\!+\!t_{ri}\!-\!2t_i) 
\end{align}
where $\|x\|_{\Psi}=x\T\Psi\inv x$ is the Mahanalobis norm.  The optimal duration $\delta_i$ can also be solved for,
\begin{align}
\delta_i &= \textstyle \frac{1}{4\nchooser{n-i+1}{2}}\Bigl(\sqrt{4\nchooser{n-i+1}{2}\norm{\hy_{\rho_{li}} \!-\! \hy_{\rho_{ri}}}^2_\Lambda \!+\!D^2}- D\Bigr) 
- \frac 1 2 (v_{\rho_{li}} \!+\! v_{\rho_{ri}} \!+\! t_{li} \!+\! t_{ri} \!-\! 2t_{i-1})
\label{eq:optdelta}
\end{align}
where $D$ is the dimensionality.  The message at the newly coalesced point has mean and covariance:
\begin{equation} \textstyle
v_{\rho_i} = \bigl(
(v_{\rho_{li}}+t_{li}-t_i)\inv +
(v_{\rho_{ri}}+t_{ri}-t_i)\inv\bigr)\inv
;
\hy_{\rho_i} = \bigl(
\frac {\hy_{\rho_{li}}}{v_{\rho_{li}}+t_{li}-t_i}+
\frac {\hy_{\rho_{ri}}}{v_{\rho_{ri}}+t_{ri}-t_i}
\bigr)v_{\rho_i}
\end{equation}
\halcomment{
\begin{align}
v_{\rho_i} &= \left(
(v_{\rho_{li}}\!+\!t_{li}-t_i)\inv +
(v_{\rho_{ri}}\!+\!t_{ri}-t_i)\inv\right)\inv \\
\hy_{\rho_i} &= \left(
\hy_{\rho_{li}}(v_{\rho_{li}}\!+\!t_{li}\!-\!t_i)\inv+
\hy_{\rho_{ri}}(v_{\rho_{ri}}\!+\!t_{ri}\!-\!t_i)\inv 
\right)v_{\rho_i} \nonumber
\end{align}
}

\halcomment{
\subsection{Example: binary vectors}

In the case of binary vectors, because each entry evolves independently, we can represent each entry separately, and the messages are just $2H$ numbers, $M_{\rho_i}^{h0}$ and $M_{\rho_i}^{h1}$ for $h=1,\ldots,H$.  The local likelihood for the $h$th entry is
\begin{align}
Z_{\rho_i}^h(\vx,\theta_i) =&
(1\!-\!q_h)M_{\rho_{li}}^{h0}M_{\rho_{ri}}^{h0} +
q_hM_{\rho_{li}}^{h1}M_{\rho_{ri}}^{h1} \nonumber \\
 &-(1\!-\!e^{-\lambda_h(t_{li}+t_{ri}-2t_i)})q_h(1\!-\!q_h)
(M_{\rho_{li}}^{h1}\!-\!M_{\rho_{li}}^{h0})
(M_{\rho_{ri}}^{h1}\!-\!M_{\rho_{ri}}^{h0})
\end{align}
We normalize each dimension independently by dividing $M_{\rho_i}^{h0}$ and $M_{\rho_i}^{h1}$ by $Z_{\rho_i}^h(\vx,\theta_i)$.  We also normalize the initial messages on the leaves similarly with
\begin{align}
Z_i^h(x_i) = (1\!-\!q_h)M_{\{i\}}^{h0} + q_hM_{\{i\}}^{h1}
\end{align}
so that the messages from leaves and coalescent points are comparable.  The optimal $\delta_i$ cannot be found analytically; we use Newton steps to solve for $\delta_i$ efficiently.  
}

\textbf{Multinomial vectors}.  Consider a Markov process acting on multinomial vectors with each entry taking one of $K$ values and evolving independently.  Entry $d$ evolves at rate $\lambda_d$ and has equilibrium distribution vector $q_d$.  The transition rate matrix is $Q_d=\lambda_d(q_h\T\bone_K-I_k)$ where $\bone_K$ is a vector of $K$ ones and $I_K$ is identity matrix of size $K$, while the transition probability matrix for entry $d$ in a time interval of length $t$ is $e^{Q_dt}=e^{-\lambda_dt}I_K+(1-e^{-\lambda_dt})q_d\T\bone_K$.  Representing the message for entry $d$ from $\rho_i$ to its parent as a vector $M_{\rho_i}^d=[M_{\rho_i}^{d1},\ldots,M_{\rho_i}^{dK}]\T$, normalized so that $q_d\cdot M_{\rho_i}^d=1$, the local likelihood terms and messages are computed as,
\begin{align}
Z_{\rho_i}^d(\vx,\theta_i) &= 
1-e^{\lambda_h(2t_i-t_{li}-t_{ri})}
\bigl(1-\textstyle \sum_{k=1}^K q_{dk}M_{\rho_{li}}^{dk}M_{\rho_{ri}}^{dk}\bigr) \\
M_{\rho_i}^d &= 
(1-e^{\lambda_d(t_i-t_{li})}(1-M_{\rho_{li}}^d))
(1-e^{\lambda_d(t_i-t_{ri})}(1-M_{\rho_{ri}}^d))/
Z_{\rho_i}^d(\vx,\theta_i)
\end{align}
Unfortunately the optimal $\delta_i$ cannot be solved analytically and we use Newton steps to compute it.
\halcomment{
\begin{align}
Q_h &= 
\lambda_h\left( \begin{bmatrix}
q_{h1} & q_{h2} & \cdots & q_{hK} \\
\vdots & \vdots &        & \vdots \\
q_{h1} & q_{h2} & \cdots & q_{hK}
\end{bmatrix} - I_K\right) \\
e^{Q_ht} &= 
e^{-\lambda_ht}I_K + (1-e^{-\lambda_ht})
\begin{bmatrix}
q_{h1} & q_{h2} & \cdots & q_{hK} \\
\vdots & \vdots &        & \vdots \\
q_{h1} & q_{h2} & \cdots & q_{hK}
\end{bmatrix} \\
N_{\rho_i}^h &= 
e^{-\lambda_ht} M_{\rho_i}^h + (1-e^{-\lambda_ht})
\begin{bmatrix}
q_h \cdot M_{\rho_i}^h \\
\vdots \\
q_h \cdot M_{\rho_i}^h
\end{bmatrix} \nonumber \\
&= 
e^{-\lambda_ht} M_{\rho_i}^h + (1-e^{-\lambda_ht}) \\
Z_{\rho_i}(\vx,\theta_i) &=
\left(
e^{-\lambda_h(t_{li}+t_{ri}-2t_i)}
\sum_{k=1}^K q_{hk}M_{\rho_{li}}^{hk}M_{\rho_{ri}}^{hk}
\right)
+
(1-e^{-\lambda_h(t_{li}+t_{ri}-2t_i)})
\left( q_h\cdot M_{\rho_{li}}^h \right)
\left( q_h\cdot M_{\rho_{ri}}^h \right) \nonumber \\
&=
\left(
e^{-\lambda_h(t_{li}+t_{ri}-2t_i)}
\sum_{k=1}^K q_{hk}M_{\rho_{li}}^{hk}M_{\rho_{ri}}^{hk}
\right)
+
(1-e^{-\lambda_h(t_{li}+t_{ri}-2t_i)})
\end{align}
Since we are assuming that the messages are normalized so that $q_h\cdot M_{\rho_i}^h=1$.
}

\subsection{Hyperparameter estimation and predictive density}

We perform hyperparameter estimation by iterating between estimating a geneology, then re-estimating the hyperparamters conditioned on this tree.  Space precludes a detailed discussion of the algorithms we use; they can be found in the supplemental material.  In the Brownian case, we place an inverse Wishart prior on $\La$ and the MAP posterior $\hat \La$ is available in a standard closed form.  In the multinomial case, the updates are not available analytically and must be solved iteratively.

Given a tree and a new individual $y'$ we wish to know: (a) where $y'$ might coalescent and (b) what the density is at $y'$.  In the supplemental material, we show that the probability that $y'$ merges at time $t$ with a given sibling is available in closed form for the Brownian motion case.  To obtain the density, we sum over all possible siblings and integrate out $t$ by drawing equally spaced samples.

\section{Experiments}

\paragraph{Synthetic Data Sets}
\begin{figure}
\includegraphics[width=\textwidth]{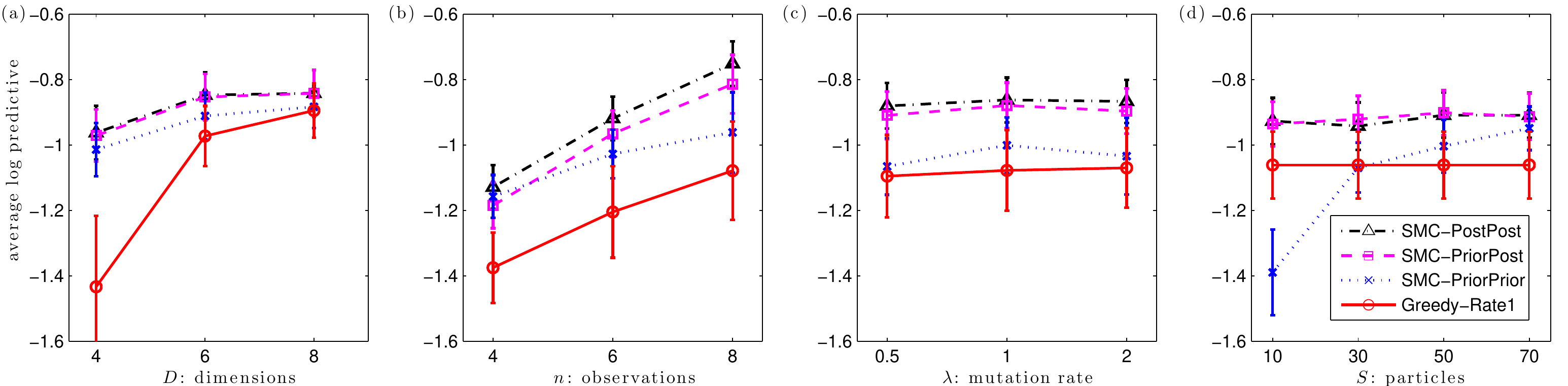}
\caption{Predictive performance of algorithms as we vary (a) the numbers of dimensions $D$, (b) observations $n$, (c) the mutation rate $\lambda$ ($\Lambda=\lambda I_D$), and (d) number of samples $S$.  In each panel other parameters are fixed to their middle values (we used $S=50$) in other panels, and we report log predictive probabilities on one unobserved entry, averaged over 100 runs.}
\label{fig:syn}
\end{figure}

In \figref{syn} we compare the various SMC algorithms and \alg{Greedy-Rate1}\footnote{We found in unreported experiments that the greedy algorithms worked about equally well.} on a range of synthetic data sets drawn from the Brownian diffusion coalescent process itself ($\Lambda=I_D$) to investigate the effects of various parameters on the efficacy of the algorithms.  
Generally \alg{SMC-PostPost} performed best, followed by \alg{SMC-PriorPost}, \alg{SMC-PriorPrior} and \alg{Greedy-Rate1}.  With increasing $D$ the amount of data given to the algorithms increases and all algorithms do better, especially \alg{Greedy-Rate1}.  This is because the posterior becomes concentrated and the \alg{Greedy-Rate1} approximation corresponds well with the posterior.  As $n$ increases, the amount of data increases as well and all algorithms perform better\footnote{Each panel was generated from independent runs.  Data set variance affected all algorithms, varying overall performance across panels.  However, trends in each panel are still valid, as they are based on the same data.}.  
However, the posterior space also increases and \alg{SMC-PriorPrior} which simply samples from the prior over genealogies does not improve as much.  We see this effect as well when $S$ is small.  As $S$ increases all SMC algorithms improve.  Finally, the algorithms were surprisingly robust when there is mismatch between the generated data sets' $\lambda$ and the $\lambda$ used by the model.  We expected all models to perform worse with \alg{SMC-PostPost} best able to maintain its performance (though this is possibly due to our experimental setup).

\paragraph{MNIST and SPAMBASE}

\begin{table}
\begin{center}
\small
\begin{tabular}{|l|c@{\hspace*{1.7ex}}c@{\hspace*{1.7ex}}c|c@{\hspace*{1.7ex}}c@{\hspace*{1.7ex}}c|}
\hline
& & {\bf MNIST} & & & {\bf SPAMBASE} & \\
& {\bf Avg-link} & {\bf BHC} & {\bf Coalescent} & {\bf Avg-link} & {\bf BHC} & {\bf Coalescent}\\
\hline
Purity    & 
$.363 \!\pm\! .004$ & 
$.392 \!\pm\! .006$ & 
$\mathbf{.412 \!\pm\! .006}$ &
$.616 \!\pm\! .007$ & 
$\mathbf{.711 \!\pm\! .010}$ & 
$.689 \!\pm\! .008$ \\
Subtree &
$.581 \!\pm\! .005$ &
$.579 \!\pm\! .005$ & 
$\mathbf{.610 \!\pm\! .005}$ &
$.607 \!\pm\! .011$ &
$.549 \!\pm\! .015$ & 
$\mathbf{.661 \!\pm\! .012}$ \\
LOO-acc & 
$.755 \!\pm\! .005$ &
$.763 \!\pm\! .005$ & 
$\mathbf{.773 \!\pm\! .005}$ &
$.846 \!\pm\! .010$ &
$.832 \!\pm\! .010$ & 
$\mathbf{.861 \!\pm\! .008}$ \\
\hline
\end{tabular}
\end{center}
\caption{Comparative results.  Numbers are averages and standard errors over 50 and 20 repeats.}
\label{tab:mnist}
\end{table}

We compare the performance of our approach (\alg{Greedy-Rate1} with 10 iterations of hyperparameter update) to two other hierarchical clustering algorithms: average-link agglomerative clustering and Bayesian hierarchical clustering \cite{HelGha2005}.  In MNIST, We use 10 digits from the MNIST data set, 20 examplars for each digit and 20 dimensions (reduced via PCA), repeating the experiment 50 times.  In SPAMBASE, we use 100 examples of 57 attributes each from 2 classes, repeating 20 times. We present purity  scores \cite{HelGha2005}, subtree scores ($\#\{\text{interior nodes with all leaves of same class}\}/(n-\#\text{classes})$) and leave-one-out accuracies (all scores between 0 and 1, higher better).  The results are in Table~\ref{tab:mnist}; as we can see, except for purity on SPAMBASE, ours gives the best performance.  Experiments not presented here show that all greedy algorithms perform about the same and that performance improves with hyperparameter updates.


\halcomment{
In Figure~\ref{fig:smc}, we compare how good the various proposals are at getting representative trees from the posterior by having them calculate a quantity proportional to the marginal likelihood.
}

\paragraph{Phylolinguistics}

\begin{figure}[t]
\subfigure[Coalescent for a subset of Indo-European languages from WALS.]{\framebox{\includegraphics[width=6.2cm]{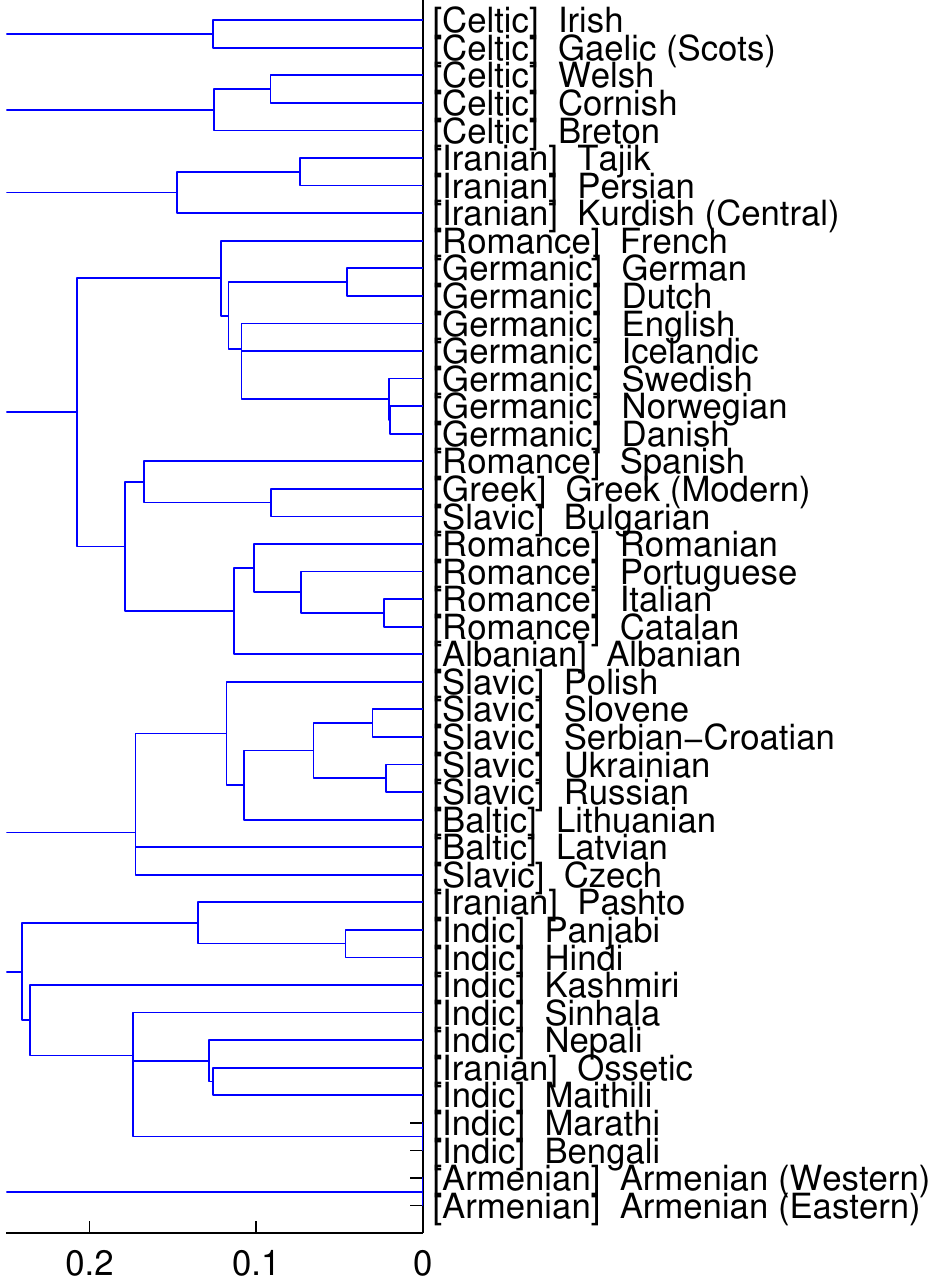}}}
\hspace{1cm}
\subfigure{
\begin{minipage}[t]{6.5cm}
\vspace{-8.5cm}
\includegraphics[width=6.5cm]{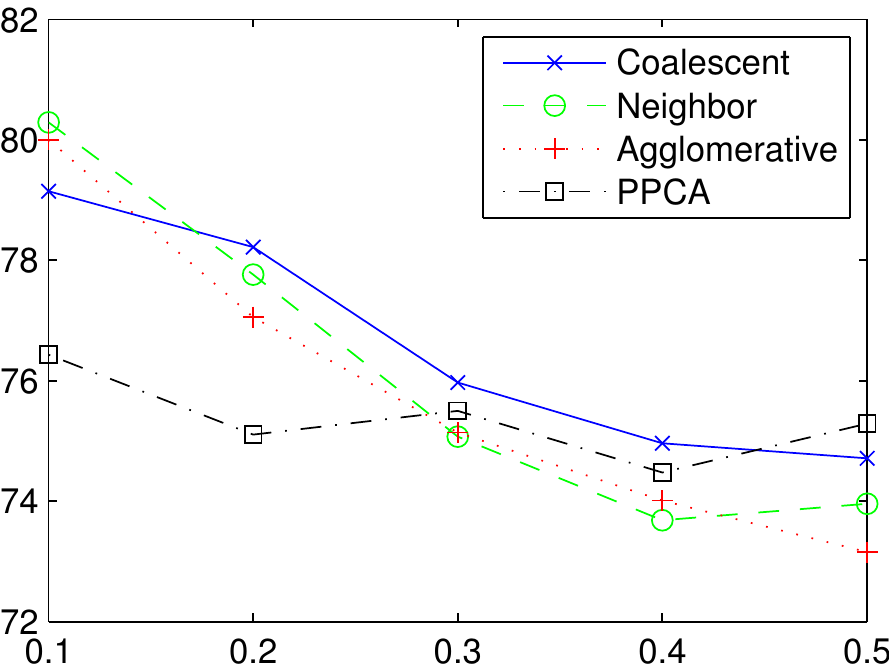}\\
{\small (b) Data restoration on WALS.  Y-axis is accuracy; X-axis is percentage of data set used in experiments.  At $10\%$, there are $N=215$ languages, $H=14$ features and $p=94\%$ observed data; at $20\%$, $N=430$, $H=28$ and $p=80\%$; at $30\%$: $N=645$, $H=42$ and $p=66\%$; at $40\%$: $N=860$, $H=56$ and $p=53\%$; at $50\%$: $N=1075$, $H=70$ and $p=43\%$.  Results are averaged over five folds with a different $5\%$ hidden each time.  (We also tried a ``mode'' prediction, but its performance is in the 60\% range in all cases, and is not depicted.)}
\end{minipage}
}
\vspace*{-1em}
\caption{Results of the phylolinguistics experiments.}
\label{fig:wals}
\end{figure}

\halcomment{
\begin{wrapfigure}{r}{6.2cm}
\vspace{-15mm}
\center\framebox{
\includegraphics[width=5.8cm]{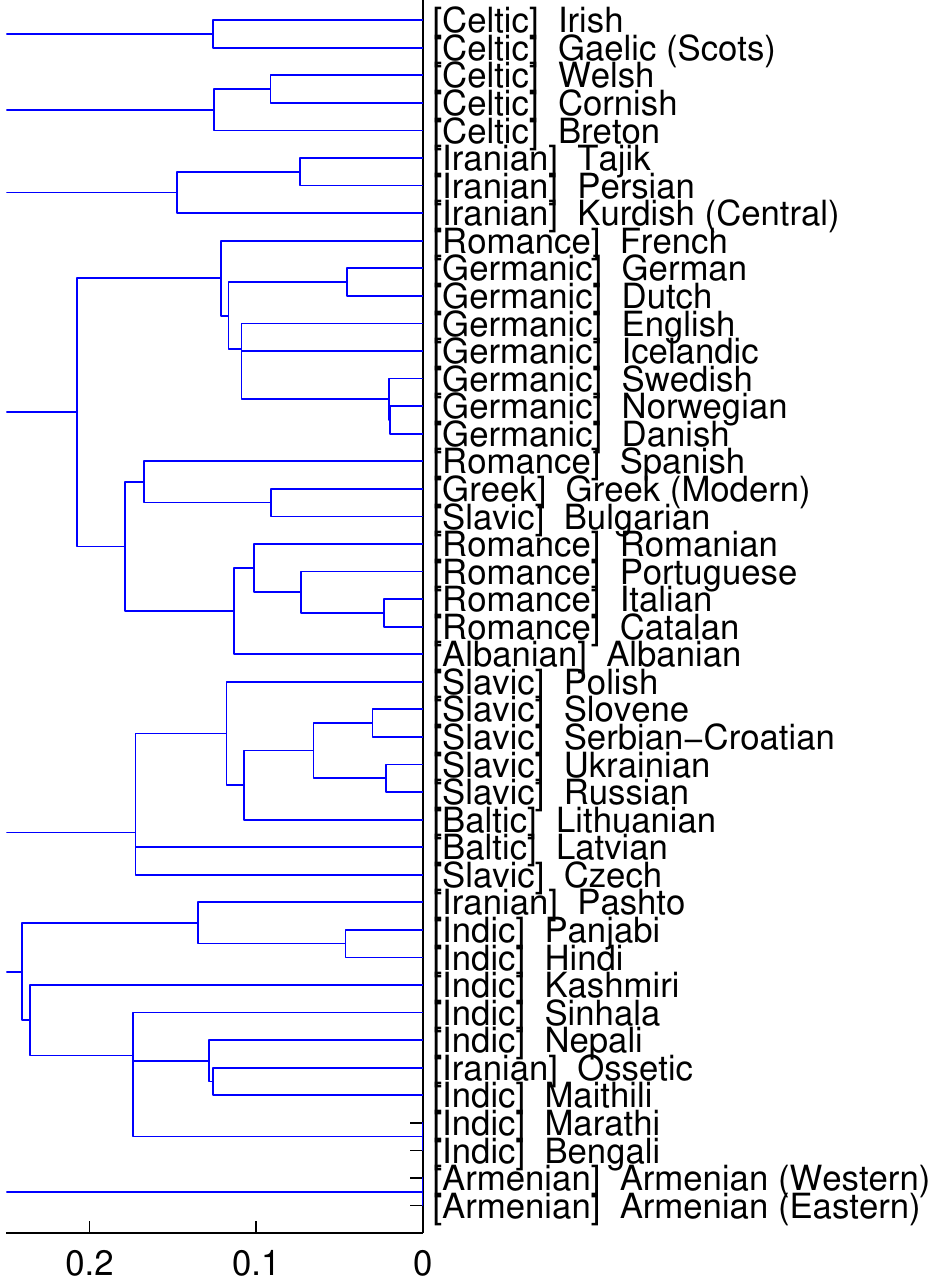}
}
\caption{Coalescent for a subset of Indo-European languages from WALS.}
\label{fig:ie-new}
\end{wrapfigure}
}
We apply our approach (\alg{Greedy-Rate1}) to a phylolinguistic problem: language evolution.  \halcomment{The majority of work on phylolinguistics has been by hand.  Namely, linguists study both historical documentation as well as language structures and attempt to identify the relationships amongst them.}  Unlike previous research \cite{mcmahon05language} which studies only phonological data, we use a full typological database of $139$ binary features over $2150$ languages: the \emph{World Atlas of Language Structures} (henceforth, ``WALS'') \cite{wals}.  \halcomment{There are $139$ \emph{features} in this database, broken down into categories such as ``Nominal Categories,'' ``Simple Clauses,'' ``Phonology,'' ``Word Order,'' etc.}  The data is \emph{sparse}: about $84\%$ of the entries are unknown.  \halcomment{many language/feature pairs, the feature value is unknown.  In fact, only about $16\%$ of all possible language/feature pairs are known.}  We use the same version of the database as extracted by \cite{daume07implication}.  Based on the Indo-European subset of this data for which at most 30 features are unknown (48 language total), we recover the coalescent tree shown in Figure~\ref{fig:wals}(a).  Each language is shown with its genus, allowing us to observe that it teases apart Germanic and Romance languages, but makes a few errors with respect to Iranian and Greek.  (In the supplemental material, we report results applied to a wider range of languages.)

Next, we compare predictive abilities to other algorithms.  We take a subset of WALS and tested on 5\% of withheld entries, restoring these with various techniques: \alg{Greedy-Rate1}; nearest neighbors (use value from nearest observed neighbor); average-linkage (nearest neighbor in the tree); and probabilistic PCA (latent dimensions in $5,10,20,40$, chosen optimistically).  We use five subsets of the WALS database of varying size, obtained by sorting both the languages and features of the database according to how many cells are observed.  \halcomment{Thus, the ``upper left'' of the matrix is dense and the ``lower right'' is sparse.}  We then use a varying percentage ($10\%-50\%$) of the densest portion.  The results are in Figure~\ref{fig:wals}(b).  The performance of PPCA is steady around 76\%.  The performance of the other algorithms degrades as the sparsity incrases.  Our approach performs at least as well as all the other techniques, except at the two extremes.

\paragraph{NIPS}


\begin{table}[t]
\footnotesize
\begin{tabular}{|@{ }c@{ }c@{ }|@{ }l@{ }|@{ }l@{ }|}
\hline
{\bf LLR} & {\bf \emph{(t)}} & {\bf Top Words} & {\bf Top Authors} \\
\hline
32.7 & \emph{(-2.71)}      & {bifurcation attractors hopfield network saddle}
     & {Mjolsness (9) Saad (9) Ruppin (8) Coolen (7)} \\
\hline
0.106& \emph{(-3.77)}     & {voltage model cells neurons neuron}
     & {Koch (30) Sejnowski (22) Bower (11) Dayan (10)} \\
\hline
83.8 & \emph{(-2.02)}    & {chip circuit voltage vlsi transistor}
     & {Koch (12) Alspector (6) Lazzaro (6) Murray (6)} \\
\hline
140.0& \emph{(-2.43)}    & {spike ocular cells firing stimulus}
     & {Sejnowski (22) Koch (18) Bower (11) Dayan (10)} \\
\hline
2.48 & \emph{(-3.66)}    & {data model learning algorithm training}
     & {Jordan (17) Hinton (16) Williams (14) Tresp (13)} \\
\hline
31.3 & \emph{(-2.76)}    & {infomax image ica images kurtosis}
     & {Hinton (12) Sejnowski (10) Amari (7) Zemel (7)} \\
\hline
31.6 & \emph{(-2.83)}    & {data training regression learning model}
     & {Jordan (16) Tresp (13) Smola (11) Moody (10)} \\
\hline
39.5 & \emph{(-2.46)}    & {critic policy reinforcement agent controller}
     & {Singh (15) Barto (10) Sutton (8) Sanger (7)} \\
\hline
23.0 & \emph{(-3.03)}    & {network training units hidden input}
     & {Mozer (14) Lippmann (11) Giles (10) Bengio (9)} \\
\hline
\end{tabular}
\caption{Nine clusters discovered in NIPS abstracts data.}
\label{tab:nipsclusters}
\end{table}

We applied \alg{Greedy-Rate1} to all NIPS abstracts through NIPS12 (1740, total).  The data was preprocessed so that only words occuring in at least 100 abstracts were retained.  The word counts were then converted to binary.  We performed one iteration of hyperparameter re-estimation.  In the supplemental material, we depict the top levels of the coalescent tree.\halcomment{top portion of the coalescent tree for the NIPS abstracts.  Each ``leaf'' is represented by the most frequent \emph{authors} under that leaf (with associated paper counts) and the most representative words in all documents under that leaf (sorted by tf-idf).  Note that author information was \emph{not} used in constructing the tree.  As we can see, there are very few authors who span multiple nodes.  For instance, Terry Sejnowski appears both in the top leaf (roughly the ``neural networks'' leaf) and the fifth leaf (the ``visual cortex'' leaf).}  Here, we use use the tree to generate a flat clustering.  To do so, we use the log likelihood ratio at each branch in the coalescent to determine if a split should occur.  If the log likelihood ratio is greater than zero, we break the branch; otherwise, we recurse down.  On the NIPS abstracts, this leads to nine clusters, depicted in Table~\ref{tab:nipsclusters}.  Note that clusters two and three are quite similar---had we used a slighly higher log likelihood ratio, they would have been merged (the LLR for cluster 2 was only $0.105$).  Note that the clustering is able to tease apart Bayesian learning (cluster 5) and non-bayesian learning (cluster 7)---both of which have Mike Jordan as their top author!

\section{Discussion}\label{sec:discussion}

We described a new model for Bayesian agglomerative clustering.  We used Kingman's coalescent as our prior over trees, and derived efficient and easily implementable greedy and SMC inference algorithms for the model.  We showed empirically that our model gives better performance than other agglomerative clustering algorithms, and gives good results on applications to document modeling and phylolinguistics.

Our model is most similar in spirit to the Dirichlet diffusion tree of \cite{Nea2001}.  Both use infinitely exchangeable priors over trees.  While \cite{Nea2001} uses a fragmentation process for trees, our prior uses the reverse---a coalescent process instead.  This allows us to develop simpler inference algorithms than those in \cite{Nea2001}, though it will be interesting to consider the possibility of developing analogous algorithms for \cite{Nea2001}.  \cite{Wil2000} also describes a hierarchical clustering model involving a prior over trees, but his prior is not infinitely exchangeable.  \cite{RoyKemMan2007} uses tree-consistent partitions to model relational data; it would be interesting to apply our approach to their setting.  Another related work is the Bayesian hierarchical clustering of \cite{HelGha2005}, which uses an agglomerative procedure returning a tree structured approximate posterior for a Dirichlet process mixture model. As opposed to our work \cite{HelGha2005} uses a flat mixture model and does not have a notion of distributions over trees.  

There are a number of unresolved issues with our work.  Firstly, our algorithms take $O(n^3)$ computation time, except for \alg{Greedy-Rate1} which takes $O(n^2)$ time.  Among the greedy algorithms we see that there are no discernible differences in quality of approximation thus we recommend \alg{Greedy-Rate1}.  It would be interesting to develop SMC algorithms with $O(n^2)$ runtime.  Secondly, there are unanswered statistical questions.  For example, since our prior is infinitely exchangeable, by de Finetti's theorem there is an underlying random distribution for which our observations are i.i.d.\ draws.  What is this underlying random distribution, and how do samples from this distribution look like?  We know the answer for at least a simple case: if the Markov process is a mutation process with mutation rate $\alpha/2$ and new states are drawn i.i.d.\ from a base distribution $H$, then the induced distribution is a Dirichlet process DP$(\alpha,H)$ \cite{Kin1982a}.  Another issue is that of consistency---does the posterior over random distributions converge to the true distribution as the number of observations grows?  Finally, it would be interesting to generalize our approach to varying mutation rates, and to non-binary trees by using generalizations to Kingman's coalescent called $\Lambda$-coalescents \cite{Pit1999}.

\small
\bibliography{refs2}
\bibliographystyle{unsrt}






\end{document}